\documentclass[letterpaper,10pt,conference]{ieeeconf}

\usepackage{amsmath}
\usepackage{amssymb}
\usepackage{array}
\usepackage{booktabs}
\usepackage{graphicx}
\usepackage{capt-of}
\usepackage{flafter}
\usepackage{stfloats}
\usepackage{flushend}
\usepackage[hidelinks]{hyperref}

\hypersetup{
    pdftitle={From Hitter to Player: Separating Motor Execution and Strategy in Competitive Humanoid Table Tennis},
    pdfauthor={Fanrong Dong, Ziyi Sun, Ziqi Han, Junhan Sun, Lei Ye, Zetong Jing, Hao Zhao, Shaoxun Liu, Rongrong Wang}
}

\title{\LARGE \bf
From Hitter to Player: Separating Motor Execution and Strategy in Competitive Humanoid Table Tennis
}

\author{%
Fanrong Dong$^{1,6,\dagger}$, Ziyi Sun$^{2,6,\dagger}$, Ziqi Han$^{3,6}$, Junhan Sun$^{1,6}$, Lei Ye$^{4,6}$,\\
Zetong Jing$^{1,6}$, Hao Zhao$^{5,6,*}$, Shaoxun Liu$^{1,*}$, Rongrong Wang$^{1}$%
\thanks{%
\raggedright
$^{1}$~Zhejiang University.\quad
$^{2}$~Ocean University of China.\quad
$^{3}$~Tongji University.\quad
$^{4}$~Shanghai Innovation Institute.\quad
$^{5}$~Tsinghua University.\quad
$^{6}$~RoboParty Lab.\quad
$^{\dagger}$~These authors contributed equally.\quad
$^{*}$~Corresponding authors.%
}}

\IEEEoverridecommandlockouts
\IEEEaftertitletext{%
\vspace{-1\baselineskip}%
\begin{minipage}{\textwidth}
    \centering
    {\small Project page: \href{https://mufanmufan.github.io/hitter-to-player/}{https://mufanmufan.github.io/hitter-to-player/}}\\[3pt]
    \includegraphics[width=0.94\textwidth,trim=0 2bp 48bp 5bp,clip]{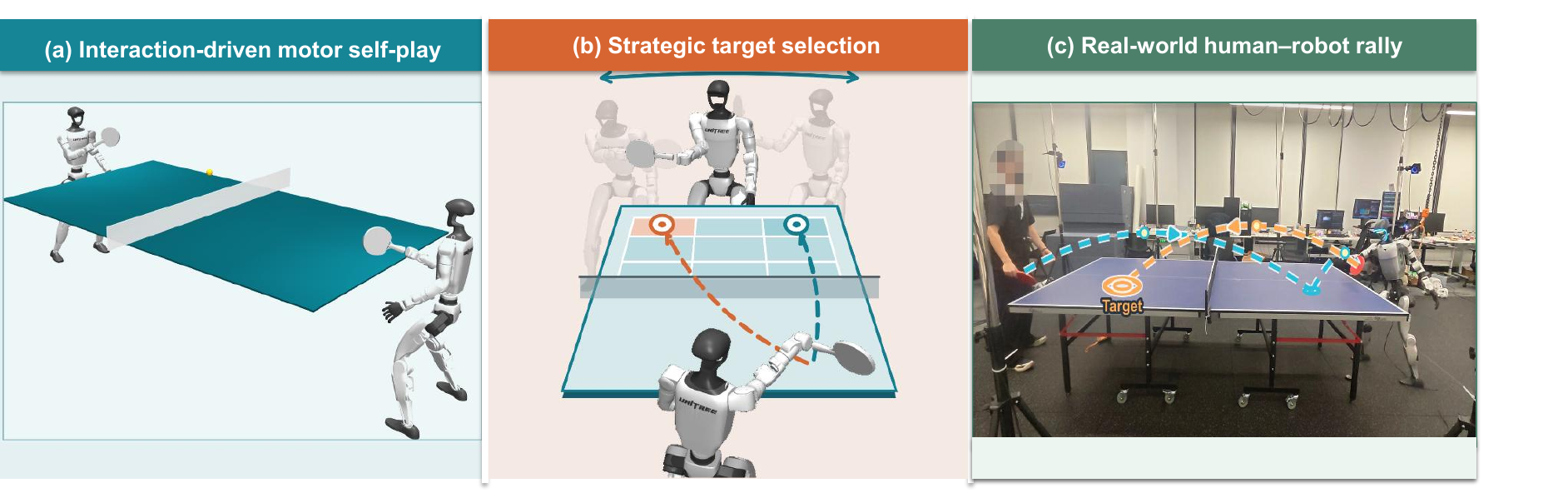}
    \setlength{\abovecaptionskip}{2pt}
    \captionof{figure}{Humanoid table-tennis training and deployment. (a) Dual-humanoid self-play generates closed-loop rally states for motor adaptation. (b) A strategic policy selects landing targets during simulated rallies. (c) The resulting hierarchy is deployed on a real G1 for human--robot play.}
    \label{fig:capability}
\end{minipage}}

\begin{document}

\bstctlcite{IEEEtranBSTCTL}

\maketitle
\thispagestyle{empty}
\pagestyle{empty}

\begin{abstract}
Competitive humanoid table tennis requires two distinct abilities: executing a desired return and deciding where to return the ball. We separate these two problems, with the ball landing target serving as an interface between motor control and strategy. (1) We first train a whole-body motor policy conditioned on the desired landing target, which maps the target to an executable humanoid stroke through a model-based strike planner and a learned controller. (2) We then improve the motor policy through self-play in closed-loop rallies, where the robot learns to handle incoming balls generated by another player rather than only prescribed incoming balls. At this stage, the landing targets are still sampled without considering the opponent, so the policy improves its execution without learning game strategy. (3) Finally, we freeze the entire motor stack and train a stroke-level strategic policy to choose landing targets through competitive self-play.

Motor self-play improves both landing accuracy and legal-return rate during rallies. More importantly, with the entire motor stack fixed, changing only the target-selection policy leads to large differences in game outcomes: the learned strategy wins 89\%, 82\%, and 73\% of points against center, random, and away-from-opponent baselines, respectively. The learned strategy also changes where it sends the ball according to the opponent's position and the previous target, showing that self-play learns more than simply aiming at fixed ``good'' locations. On hardware, the same frozen execution stack and strategic interface support complete 11-point games against a recreational human opponent, including multi-stroke rallies with both forehand and backhand returns.
\end{abstract}

\section{Introduction}

Robotic table tennis now spans precise target-conditioned returns, competitive play, and whole-body execution on legged robots~\cite{ding2022precision,dambrosio2025competitive,nguyen2025spinaware}. Recent humanoid systems coordinate striking and footwork during physical returns~\cite{su2025hitter,hu2025pace,ren2026smash}, while hierarchical and competitive systems study game-level decisions on different execution platforms~\cite{mahjourian2018selfplay,dambrosio2025competitive}. These advances motivate our question: how can a whole-body humanoid adapt execution to interaction-generated rally states and learn target selection through a shared command that separates motor from strategic adaptation?

Competitive play couples where to return with the ability to realize that return. A target can be accurate yet easy for the opponent to reach, or tactically promising yet difficult for the robot to execute. Adapting selection and execution simultaneously obscures their contributions to game outcomes. A continuous landing command supplies a control and experimental boundary: the executor learns to realize the command, and the selector learns which command to request.

Here, strategic decision making refers specifically to selecting the landing target of each return from the current rally state, whereas motor execution realizes that command through strike planning and whole-body control. As a task-space outcome, the landing target retains the same semantics as motor parameters change, providing a common variable for motor adaptation, strategic learning, and controlled evaluation. Freezing the executor before strategic learning then allows competitive differences to be evaluated under fixed learned motor capability.

This separation also gives motor refinement a distinct role. Prescribed balls provide broad initial striking experience, whereas closed-loop rallies recursively generate incoming states through both players' realized returns. We therefore adapt the executor to these interaction-generated states before training strategic selection. Opponent-independent target sampling preserves the landing-target semantics during this adaptation. Thus, Stage~II changes the execution-state distribution without opponent-conditioned tactical information; this is introduced only after freezing the executor in Stage~III.

\begin{figure*}[!t]
    \centering
    \includegraphics[width=\textwidth]{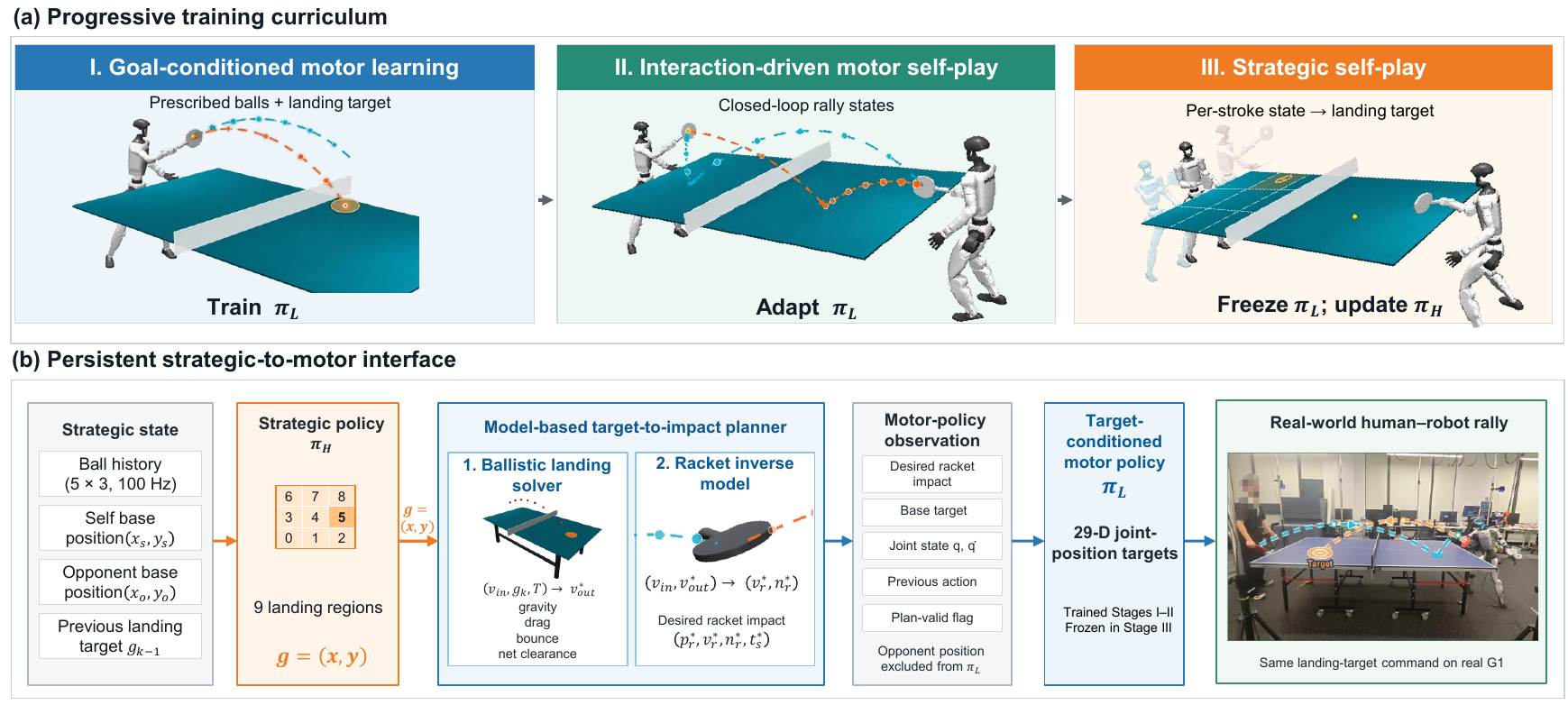}
    \caption{Progressive training and the persistent strategic-to-motor interface. (a) Target-conditioned motor training is followed by interaction-driven motor self-play; the resulting motor layer is then frozen for strategic self-play. (b) At each stroke event, the strategic policy selects a landing region whose induced continuous target conditions both the model-based planner and learned whole-body controller. Opponent position enters only the strategic policy, preserving the separation between tactical selection and motor execution.}
    \label{fig:system_overview}
\end{figure*}

Figs.~\ref{fig:capability} and~\ref{fig:system_overview} illustrate the interaction settings and three-stage hierarchy. The same continuous landing-target interface connects strategic choice, model-based strike planning, and learned whole-body control throughout motor adaptation and frozen-executor strategic learning, enabling strategic comparison under matched motor capability.

Our contributions are threefold:
\begin{itemize}
    \item \textbf{Persistent target-conditioned interface.} One continuous landing command links strategy, planning, and learned whole-body execution across all three stages.
    \item \textbf{Staged motor-to-strategy self-play.} Stage II trains execution in closed-loop rallies with opponent-independent targets; Stage III freezes it for strategic selection.
    \item \textbf{Controlled strategic evaluation under matched execution.} Learned, Center, Random, and Away share the same frozen motor actor, observation normalizer, planner, and downstream controller; differences therefore reflect target selection, not motor parameters.
\end{itemize}

\section{Related Work}
\label{sec:related_work}

\subsection{Target-Conditioned Robotic Table Tennis}

Early robotic table-tennis systems decomposed returns into prediction, interception, racket-state generation, and motion, using learned ball/landing mappings, striking primitives, reachable contact selection, and trajectory optimization~\cite{matsushima2005learning,mulling2013striking,huang2015striking,koc2018trajectory}.

Data-driven approaches now learn this mapping through model-free joint control, physical-robot reinforcement learning, and sample-efficient adaptation~\cite{gao2020modelfree,buchler2022scratch,tebbe2021sample}. GoalsEye~cite{ding2022precision} learns arbitrary landing-target-conditioned control through physical practice, while a high-speed system study combines target-conditioned policies with low-latency perception, simulation, and physical evaluation~cite{dambrosio2023system}. These works realize desired returns; we connect execution to strategic target selection through learned whole-body control.

Spin-aware returns have also been studied with curriculum reinforcement learning in simulation~\cite{hu2025spin} and spin-aware whole-body model predictive control on a physical quadruped~\cite{nguyen2025spinaware}.

\subsection{Competitive Strategy and Opponent Adaptation}

Earlier table-tennis decision systems modeled opponent preferences for preparation poses~\cite{wang2011opponent} and anticipatory action under uncertain human behavior~\cite{wang2017anticipatory}; M\"ulling \emph{et al.}~\cite{mulling2014strategy} inferred player-specific strategies from demonstrations with inverse reinforcement learning. Thus opponent and interaction state inform decisions beyond single-return mechanics.

i-Sim2Real alternates simulation learning with physical interaction to update cooperative policies and human behavior models~\cite{abeyruwan2022isim2real}; it highlights the importance of encountered states, whereas our staged design separates target-execution adaptation from competitive selection.

Hierarchical learning is especially close to our setting. Mahjourian \emph{et al.}~\cite{mahjourian2018selfplay} combined a model-free strategy layer, learned ball dynamics, model-based striking, and analytic control with self-play. D'Ambrosio~\emph{et al.}~\cite{dambrosio2025competitive} paired low-level skills and capability descriptors with high-level selection and online adaptation for play against unseen humans. Ace~\cite{durr2026ace} extends this envelope with high-speed perception, learned control, and matches against elite and professional players. Together, these works establish hierarchical strategy and skill selection in robot table tennis~\cite{mahjourian2018selfplay,dambrosio2025competitive}. We instead use a persistent continuous landing target to adapt learned whole-body execution to interaction-generated rallies, freeze it, and learn selection under matched motor capability.

\subsection{Humanoid Table Tennis}

HITTER, PACE, and SMASH demonstrate whole-body execution, consecutive returns, and autonomous perception for humanoid table tennis. \mbox{Table~\ref{tab:humanoid_comparison}} summarizes these capabilities. We use a continuous landing target as a reusable execution command and learn strategic selection over it.

\begin{table}[t]
    \refstepcounter{table}
    \label{tab:humanoid_comparison}
    \noindent\parbox{\linewidth}{\footnotesize\raggedright
        \textbf{TABLE~\Roman{table}:} Capability comparison with recent humanoid table-tennis systems. Checks indicate explicitly demonstrated capabilities; dashes indicate capabilities not demonstrated in the cited system, rather than incapability.\par}
    \vspace{2pt}
    \centering
    \footnotesize
    \renewcommand{\arraystretch}{1.02}
    \setlength{\tabcolsep}{0.8pt}
    \begin{tabular}{@{}
        >{\raggedright\arraybackslash}m{0.165\linewidth}
        >{\centering\arraybackslash}m{0.10\linewidth}
        >{\centering\arraybackslash}m{0.12\linewidth}
        >{\centering\arraybackslash}m{0.12\linewidth}
        >{\centering\arraybackslash}m{0.13\linewidth}
        >{\centering\arraybackslash}m{0.18\linewidth}
        >{\centering\arraybackslash}m{0.13\linewidth}@{}}
        \toprule
        \textbf{Method} & \shortstack{\textbf{Whole-}\\\textbf{body}} & \shortstack{\textbf{Consec.}\\\textbf{returns}} & \shortstack{\textbf{Learned}\\\textbf{WBC}} & \shortstack{\textbf{Onboard}\\\textbf{vision}} & \shortstack{\textbf{Target-}\\\textbf{conditioned}\\\textbf{landing}} & \shortstack{\textbf{Learned}\\\textbf{target}\\\textbf{strategy}} \\
        \midrule
        HITTER~\cite{su2025hitter} & $\checkmark$ & $\checkmark$ & $\checkmark$ & -- & -- & -- \\
        PACE~\cite{hu2025pace} & $\checkmark$ & $\checkmark$ & $\checkmark$ & -- & -- & -- \\
        SMASH~\cite{ren2026smash} & $\checkmark$ & $\checkmark$ & $\checkmark$ & $\checkmark$ & -- & -- \\
        \textbf{Ours} & $\checkmark$ & $\checkmark$ & $\checkmark$ & -- & $\checkmark$ & $\checkmark$ \\
        \bottomrule
    \end{tabular}
\end{table}

Other humanoid work uses kinematic pose interfaces~\cite{he2024omnih2o}, residual action corrections~\cite{he2025asap}, motion tracking with guided diffusion~\cite{truong2025beyondmimic}, and human-motion data for target-directed physical tennis~\cite{zhang2026latent}. These interfaces reuse motion capability; ours expresses the task outcome, the ball's landing target, across motor refinement and strategic learning.

\subsection{Self-Play and Multi-Timescale Embodied Learning}

Multi-agent competition generates adaptive curricula and complex behavior~\cite{bansal2018competition}. Staged learning connects motor control to team decisions in simulated humanoid football~\cite{liu2022football}; physical bipedal soccer combines agile skills and opponent-sensitive tactics~\cite{haarnoja2023soccer}. Reusable adversarial embeddings support downstream simulated-character tasks~\cite{peng2022ase}, while hierarchical walking and reaching policies support whole-body benchmarks~\cite{sferrazza2024humanoidbench}. Alongside table-tennis self-play~\cite{mahjourian2018selfplay}, these results motivate reusable execution. We distinguish interaction roles: motor self-play changes encountered states under opponent-independent targets, whereas strategic self-play learns selection over a frozen layer.

\section{System Overview and Problem Formulation}
\label{sec:system_overview}

\subsection{Humanoid Table-Tennis System}

Fig.~\ref{fig:system_overview} summarizes the staged training process and the shared deployment interface. The system comprises a Unitree G1, table, racket, and external ball tracker. Ball measurements update at 100~Hz; the learned whole-body controller runs at 50~Hz and commands 29 actuated joints; strategic decisions occur once per stroke. The G1--human interface receives the human's planar position from motion capture.

\subsection{Strategic-to-Motor Interface}

Strategic states and landing targets use a canonical player-oriented table frame, so both simulated players face the same abstract task. At stroke event $k$ of the controlled player, the strategic policy $\pi_H$ maps the 21-D state $\mathbf{s}_k$ to a region $z_k\in\{0,\ldots,8\}$ and a continuous landing target $\mathbf{g}_k$ in meters:
\begin{alignat}{2}
    z_k &\sim \pi_H(\cdot\mid\mathbf{s}_k),\quad
    &\mathbf{g}_k &= \mathbf{c}_{z_k}+\boldsymbol{\epsilon}_k,\notag\\
    \mathbf{u}_t &= \mathcal{P}(\hat{\mathbf{b}}_t,\mathbf{g}_k),
    &\mathbf{a}_t &= \pi_L(\mathbf{x}_t^L,\mathbf{u}_t,\mathbf{g}_k),\notag\\
    \mathbf{q}^{\mathrm{cmd}}_t
    &= \operatorname{clip}\!\left(
      \mathbf{q}^{\mathrm{default}}+\boldsymbol{\alpha}_{a}\odot\mathbf{a}_t
    \right). &&
    \label{eq:strategic_interface}
\end{alignat}
Here $\mathbf{c}_{z_k}$ is the selected region center and $\boldsymbol{\epsilon}_k$ is bounded training jitter (Section~\ref{sec:strategic_selfplay}). The command $\mathbf{g}_k$ is held for the stroke. At motor step $t$, the fixed planner $\mathcal{P}$ uses estimated ball position and velocity $\hat{\mathbf{b}}_t$ to supply task-relative impact and base targets, time-to-strike, and validity in $\mathbf{u}_t$. The remaining robot/task inputs $\mathbf{x}_t^L$ include proprioception, ball history, and the previous action. The actor consumes $\mathbf{x}_t^L$, these planner-derived inputs, and the direct landing command $\mathbf{g}_k$, producing a dimensionless action $\mathbf{a}_t\in\mathbb{R}^{29}$. The nominal joint-target mapping applies joint-wise action scales $\boldsymbol{\alpha}_{a}$ in radians and default-pose offsets $\mathbf{q}^{\mathrm{default}}$, then clips elementwise to the configured bounds $[\mathbf{q}^{\mathrm{soft,min}},\mathbf{q}^{\mathrm{soft,max}}]$. Eq.~\eqref{eq:strategic_interface} thus connects strategic state to joint commands. Opponent position enters only $\pi_H$, preventing direct opponent-conditioned placement by $\pi_L$.

\subsection{Hierarchical Learning Problem}

Three stages share this landing-target interface.

\noindent\textbf{Stage I---Single-ball motor training.}
The motor layer trains on prescribed incoming balls and continuous targets.

\noindent\textbf{Stage II---Interaction-driven motor self-play.}
Two simulated G1 robots generate continuous rallies. The learner faces a lagged motor-policy snapshot, while landing targets remain independent of opponent position. This changes the source of incoming balls without introducing tactical selection.

\noindent\textbf{Stage III---Strategic self-play.}
The motor layer is frozen, and the strategic policy learns landing-region selection at stroke events. Both players use the same motor checkpoint. Point outcomes and execution-oriented rewards train the policy without handcrafted tactical rewards.

\section{Landing-Target-Conditioned Humanoid Motor Control}
\label{sec:motor_control}

The motor layer converts measured ball history and a continuous landing target into whole-body joint commands through a model-based strike planner and a learned controller.

\subsection{Ball State Estimation}

The planner estimates velocity by a least-squares fit to nine measured positions from a continuously sampled 100-Hz buffer, normally spanning 80~ms. The motor and strategic observations use the five latest available positions, normally spanning 40~ms, independently of the 50-Hz control rate. At reset, the available initial measurement pads the history until new samples arrive.

\subsection{Model-Based Strike Planning}

The planner propagates the estimated ball state under gravity and quadratic drag, includes table bounce when encountered, and predicts interception at a fixed strike plane. Given $\mathbf{g}_k$, it searches outgoing flight times under net-clearance and racket-speed constraints. The selected flight determines the outgoing-ball velocity; an impact model produces target racket position, velocity, and normal, with a compatible base target locked for the stroke. In Fig.~\ref{fig:system_overview}, $T$ is outgoing flight duration and $\mathbf{v}_{\mathrm{in}}$ and $\mathbf{v}_{\mathrm{out}}^*$ are incoming and desired outgoing ball velocities. The desired impact comprises racket position $\mathbf{p}_r^*$ (m), velocity $\mathbf{v}_r^*$ (m/s), unit normal $\mathbf{n}_r^*$, and planned strike time $t_s^*$ (s); motor inputs use transformed task references and time-to-strike rather than the absolute strike timestamp. Joint angles $\mathbf{q}$ and angular velocities $\dot{\mathbf{q}}$ are measured in rad and rad/s.

\subsection{Learned Whole-Body Control}

The 50-Hz actor $\pi_L$ is an observation-normalized MLP with ELU hidden layers of widths 512, 256, and 128; it maps the 130-D observation to a 29-D dimensionless joint action. A fixed action-to-joint-target mapping applies joint-wise scaling, default-pose offsets, and configured target limits; the resulting joint-position commands are tracked by joint-level PD control. The observation groups robot/task state and proprioception, recent ball positions, the previous action, planner-derived racket and base targets in task-relative representations with time-to-strike and plan validity, and the directly included landing command $\mathbf{g}_k$. Opponent position is provided only to the strategic policy and is excluded from the motor-policy observation.

For the frozen motor checkpoint used in strategic evaluation, the low-level training reward is
\begin{equation}
r_t^L=\sum_{G\in\mathcal{G}}\sum_{j\in\mathcal{J}_G}w_jr_{j,t}
      +\lambda_{\rm AMP}(n)r_t^{\rm AMP},
\label{eq:motor_reward}
\end{equation}
Here $\mathcal{G}$ comprises task, strike, body, and safety groups; $\mathcal{J}_G$ indexes terms in group $G$, $w_j$ is the configured weight, and $n$ is the training update. The groups cover landing and return outcomes, planner-relative strike and contact quality, posture and recovery, and action, joint-limit, and collision terms. The scheduled additive AMP term~\cite{peng2021amp} is training-only; no reward term for this frozen executor uses opponent position.

\subsection{Landing-Target Motor Training}

Stage~I samples continuous targets and a broad mixture of direct, endline-bounce, and sideline-bounce incoming balls. It models drag, randomizes ball mass, and supplies the same target to both planner and controller. Landing realization is evaluated independently in Section~\ref{sec:experiments}.

\section{Interaction-Driven Motor Self-Play}
\label{sec:motor_selfplay}

Stage~II preserves the motor task but replaces prescribed incoming balls with returns generated by a simulated G1 opponent.

% Declared before the supporting Stage II diagnostic so this primary
% full-width result retains Fig. 3 and leads Page 5.
\begin{figure*}[t]
    \centering
    \includegraphics[width=\textwidth]{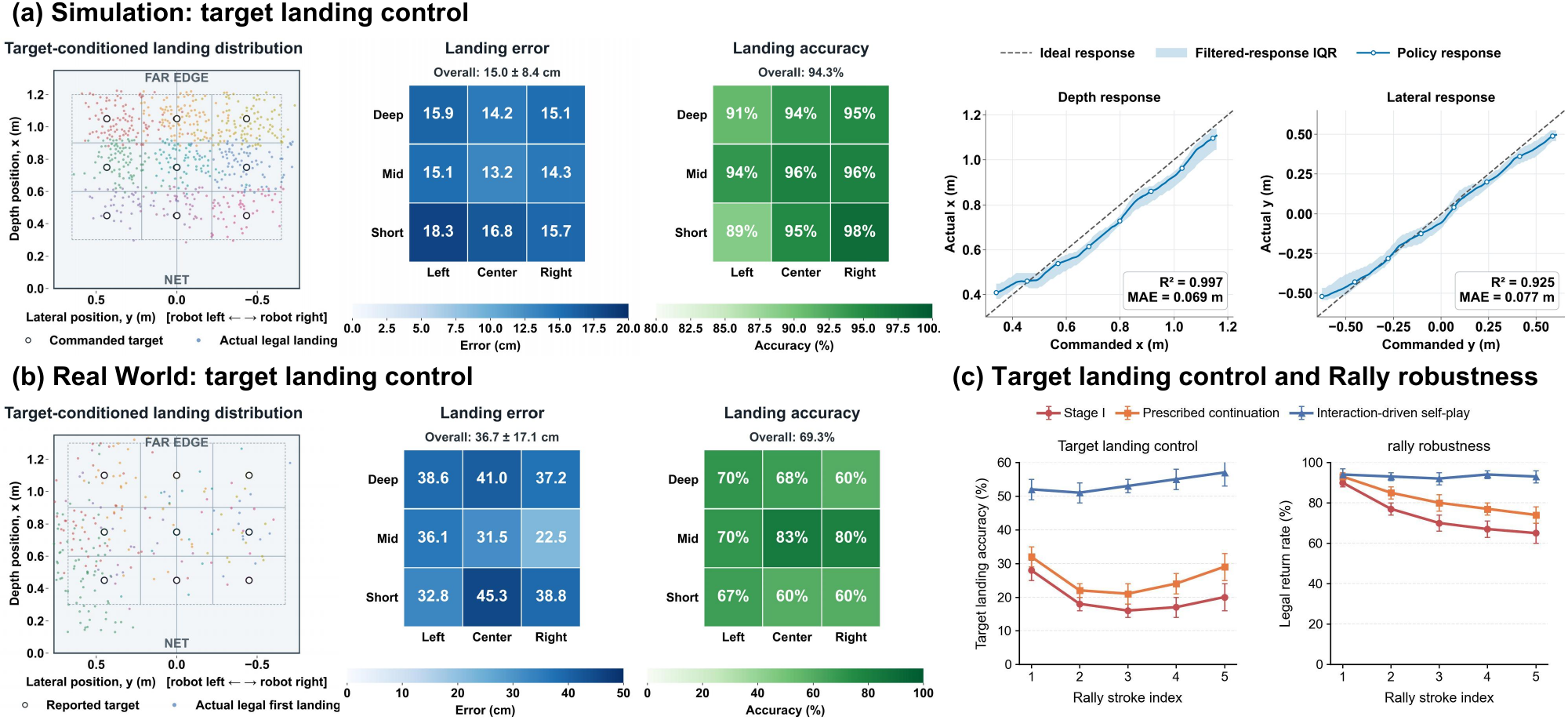}
    \caption{Target-conditioned execution in single-return and closed-loop rally evaluations. (a) Simulated landing distributions, per-target error and accuracy, and command--response curves characterize continuous landing-command realization. (b) Hardware trials evaluate the same interface across nine targets. (c) Under matched rally conditions, Stage~I is compared with matched-budget prescribed continuation and interaction-driven self-play, both initialized from the same Stage-I checkpoint and trained with the same additional budget. Prescribed continuation retains the Stage-I incoming-ball distribution, whereas interaction-driven self-play uses rally-generated incoming states. The latter achieves higher target-landing accuracy and legal-return rate across rally depth. Curves show means over eight seeds; error bars denote $\pm$1 SEM.}
    \label{fig:target_landing_control}
\end{figure*}

\subsection{Interaction-Generated Incoming States}

\begin{figure}[t]
    \centering
    \includegraphics[width=\linewidth]{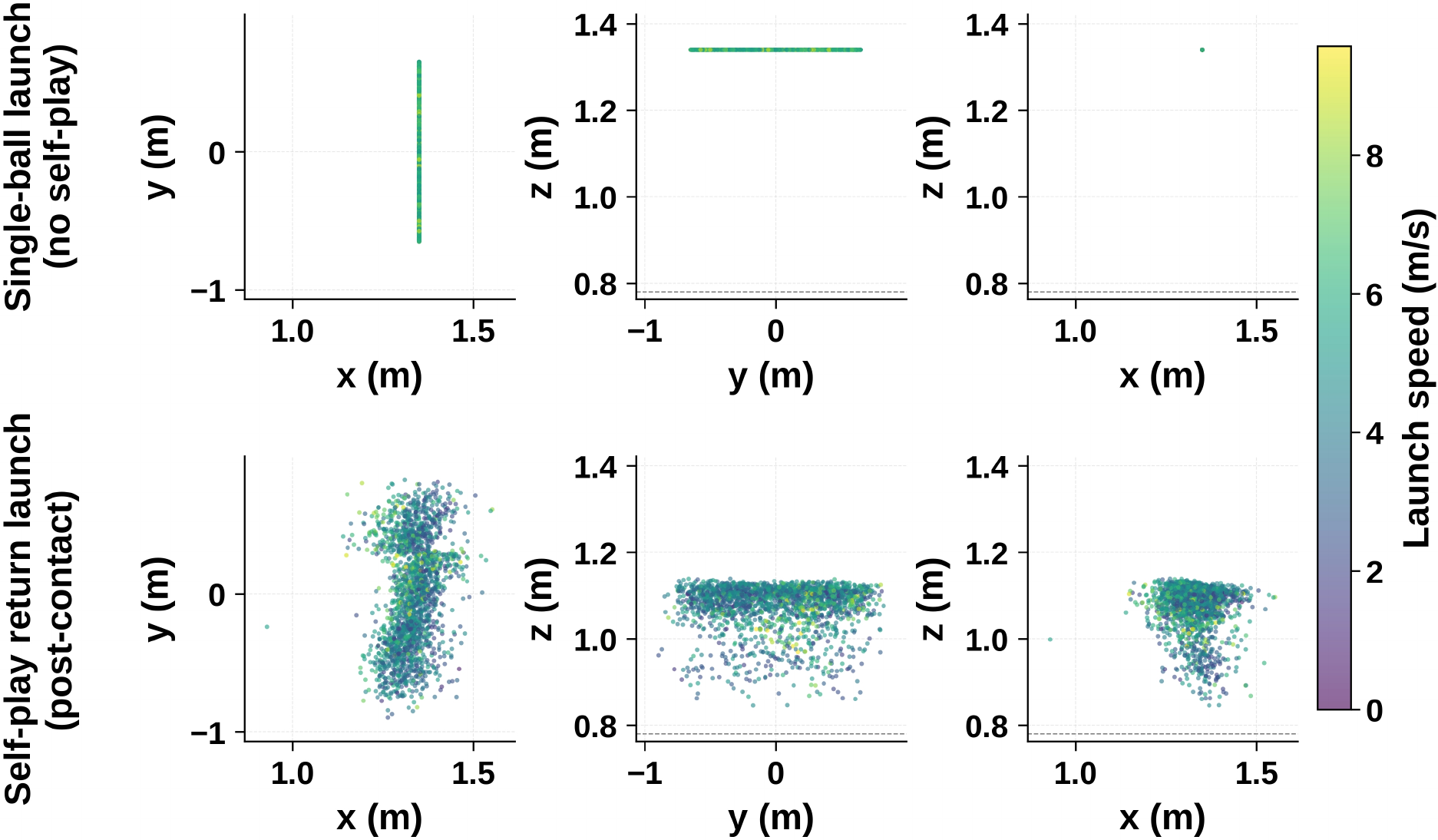}
    \caption{Launch-state geometry encountered by the motor policy. Prescribed launches (top) and motor-self-play racket returns (bottom) are shown in the receiver's canonical frame as $x$--$y$, $y$--$z$, and $x$--$z$ projections; color denotes speed. The comparison characterizes interaction-generated launch states, not strike-plane states.}
    \label{fig:launch_state_distribution}
\end{figure}

Prescribed launch states induce an incoming-ball distribution through the ball dynamics. In a rally, incoming states instead depend recursively on both players' realized strokes and errors. This change motivates adapting the motor layer to closed-loop interaction. Requested targets remain independent of opponent position, separating the source of incoming balls from tactical placement. Fig.~\ref{fig:launch_state_distribution} illustrates launch-state geometry; it is not a comparison of states at the strike plane.

\subsection{Canonical Dual-Humanoid Rally Environment}

Each player has independent dynamics, control, sensing, ball history, and planner state. Player~A uses the table frame directly; Player~B applies $R=\operatorname{diag}(-1,-1,1)$ to table-relative positions and velocities, a $180^\circ$ rotation about the table vertical axis. Both policies therefore receive near-side player coordinates and targets on the far half. A continuous rally alternates the expected hitter; a legal handover requires that player's racket contact, net crossing, and first landing on the opponent half. Illegal strokes and invalid robot states terminate the point.

\subsection{Motor Self-Play Training}

Only Player~A is optimized; Player~B uses a lagged snapshot of its motor actor and observation normalizer. Training copies both at periodic PPO-update boundaries and holds the opponent fixed between copies. Both players receive independently sampled targets, and neither observes opponent position. Stage~II retains the landing-target interface and execution task, together with the Stage~I planner, controller, and observation/action structure, while adapting to returns produced by this evolving opponent.

\section{Strategic Self-Play}
\label{sec:strategic_selfplay}

In Stage~III, the motor layer is frozen while only stroke-level landing-region selection is learned. The learning problem therefore changes target selection without updating motor parameters.

\subsection{Stroke-Level Strategic State}

At event $k$, $\pi_H$ receives $\mathbf{s}_k\in\mathbb{R}^{21}$ in the controlled player's canonical frame (Table~\ref{tab:strategic_state}). Ball history supplies recent trajectory context without an explicit velocity input; self and opponent positions supply the current interaction configuration. The previous target $\mathbf{g}_{k-1}$ records the actual command, including jitter.

\begin{table}[!ht]
    \refstepcounter{table}
    \label{tab:strategic_state}
    \noindent\parbox{\linewidth}{\footnotesize\raggedright
        \textbf{TABLE~\Roman{table}:} Stroke-level strategic state.\par}\par
    \vspace{2pt}
    \raggedright
    \footnotesize
    \renewcommand{\arraystretch}{1.02}
    \setlength{\tabcolsep}{2pt}
    \noindent\begin{tabular*}{\linewidth}{@{\extracolsep{\fill}}lcl@{}}
        \toprule
        \textbf{Input group} & \textbf{Dim.} & \textbf{Information role} \\
        \midrule
        Five ball positions & 15 & Recent trajectory \\
        Self base $xy$ & 2 & Player configuration \\
        Opponent base $xy$ & 2 & Opponent context \\
        Previous target $xy$ & 2 & Last landing command \\
        \bottomrule
    \end{tabular*}\par
\end{table}

\subsection{Strategic Landing Actions}

The nine region centers combine depths of 0.35, 0.75, and 1.10~m with lateral coordinates of $-0.45$, 0, and 0.45~m in the canonical table frame. Bounded training jitter adds up to $\pm0.05$~m in depth and $\pm0.07$~m laterally through Eq.~\eqref{eq:strategic_interface}. This grid keeps event-level selection compact while retaining a continuous motor command.

\subsection{Event-Driven Decisions and Transitions}

The initial receiver selects a target at point reset. Thereafter, selection occurs at the end of the first motor step registering the opponent's racket contact. All components of $\mathbf{s}_k$ are read at that instant: the available ball-history buffer, current self and opponent positions, and stored previous target. The ball samples can straddle contact; no future measurement or five-sample post-contact window is used. The command is held for the ensuing stroke.

Each strategic transition links a player's target choice to that same player's next decision. Its reward accounts for the executed return and the opponent's response, so outcome credit can arrive after the next decision has been triggered. A legal opponent return completes the nonterminal transition; point termination assigns the terminal outcome and ends value bootstrapping. A target selected for a stroke that never becomes legally playable receives no separate transition. Same-player successor links support generalized advantage estimation~\cite{schulman2015gae} over these irregular events.

\subsection{Self-Play Training}

The actor and critic are observation-normalized MLPs with ELU hidden layers of widths 256, 256, and 128; the actor parameterizes a nine-way categorical distribution. PPO~\cite{schulman2017ppo} uses a learning rate of $3\!\times\!10^{-4}$, $\gamma=0.99$, GAE $\lambda=0.95$, a 0.2 clipping ratio, four epochs, four minibatches, and 32{,}768 event transitions per update. Stage~III freezes the motor actor, its observation normalizer, planner, and downstream joint controller; both sides execute through the same motor checkpoint.

The curriculum randomizes learner side per point and starts with Center and Random selectors. A lagged strategic opponent is then initialized from the learner. This opponent receives no gradient updates and is refreshed from the learner every 50 PPO updates, remaining frozen between refreshes.

The strategic reward combines terminal competition with execution shaping. From the controlled player's perspective, the point component is $+1$ for a win, $-1$ for a loss, and zero at nonterminal events. A legal return adds 0.02; a legal landing inside the selected region earns a bonus that begins at 0.02 and decays linearly to zero over the first 2{,}000 PPO updates. No reward term encodes opponent distance, corners, displacement, alternation, or depth patterns; tactical placement must arise through competitive outcomes and executable target choices.

\subsection{Sim-to-Human Deployment Interface}

Deployment retains the observation, action space, planner, and frozen controller, replacing simulated opponent position with motion-captured human planar position in the same canonical frame.

\begin{figure}[t]
    \centering
    \includegraphics[width=\linewidth]{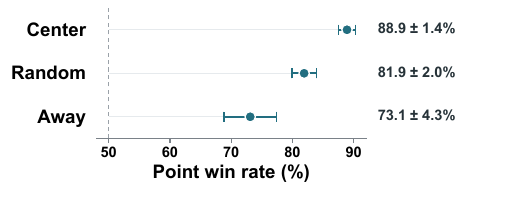}
    \caption{Competitive target selection under matched motor execution. The learned selector has higher point win rates against Center, Random, and opponent-position-aware Away. Results are mean $\pm$ sample standard deviation across eight evaluation seeds. All selectors share one frozen motor checkpoint, so differences reflect target selection under fixed learned motor parameters; the dashed line denotes 50\% equal performance.}
    \label{fig:pairwise_win_rate}
\end{figure}

\section{Experiments}
\label{sec:experiments}

Four questions assess target execution, rally refinement, matched-execution competition, and controlled target-selection sensitivity.

\subsection{Experimental Setup}

Simulation and hardware use the interfaces in Sections~\ref{sec:system_overview}--\ref{sec:motor_selfplay}. Hardware uses measured ball positions; human interaction adds motion-captured opponent position. Execution is assessed by landing distributions, command-response curves, landing accuracy, and legal-return rate; competitive selection by pairwise win rates and controlled target-selection maps.

\paragraph{Matched-budget prescribed continuation}
To separate the effect of additional optimization from interaction-generated incoming states, we continue from one Stage-I checkpoint under two matched-budget conditions. Prescribed continuation samples the Stage-I incoming distribution, whereas the interaction-driven branch uses the Stage-II self-play environment to generate incoming balls through closed-loop rallies. Both branches use identical PPO updates and environment steps, with reward definition, model-based strike planner, target sampling, network architecture, optimizer, and learning-rate schedule held fixed; both checkpoints and the original Stage-I policy are evaluated under one fixed-opponent closed-loop rally protocol.

Simulation legal landing rate counts all attempts; landing error and accuracy in Fig.~\ref{fig:target_landing_control}(a,b) count only legal first opponent-half landings. Landing accuracy uses a 30-cm tolerance in simulation and 45~cm on hardware. These domain-specific thresholds are not compared directly across domains. Hardware landing error and accuracy are conditional on legal first landings, not end-to-end return success. Fig.~\ref{fig:target_landing_control}(c) uses matched rally conditions with 30{,}000 rallies per checkpoint per seed; points are equal-weight means of eight seed-level rates with $\pm$1 SEM error bars. At each stroke index, all return opportunities reaching that stroke enter the denominator, with unsuccessful returns counted as failures.

For strategic cross-play, all selectors share one frozen motor checkpoint and the learned condition one fixed strategic checkpoint. Learned inference uses the categorical argmax; the motor actor uses its deterministic Gaussian mean. At evaluation, $\boldsymbol{\epsilon}_k=\mathbf{0}$ and selectors command exact region centers: Center selects Zone~4, $\mathbf{c}_4=(0.75,0)$~m; Random hashes the evaluation seed, environment, episode, decision index, and side to one of nine regions; Away selects the region center farthest from the current opponent. Each unordered pair is evaluated over eight rollout seeds with 200 simulated point episodes per seed, balanced across table side and initial receiver. Point win rate is computed within each seed and averaged equally across seeds; Fig.~\ref{fig:pairwise_win_rate} shows the learned comparisons.

\begin{figure}[t]
    \centering
    \includegraphics[width=\linewidth,pagebox=mediabox,viewport=222 94 704 468.7,clip]{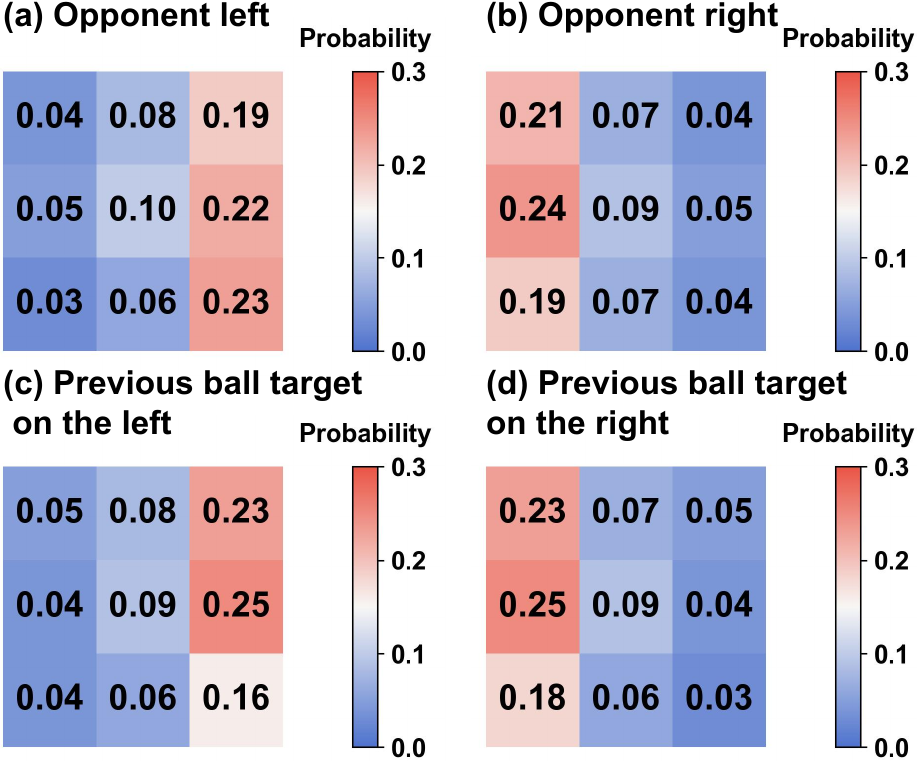}
    \caption{Controlled local sensitivity of strategic target selection. Only opponent position changes in (a,b), with previous target centered; only previous target changes in (c,d), with opponent centered. All other inputs are fixed. Each $3\times3$ map gives empirical region-selection probabilities from 500 categorical samples on a shared color scale.}
    \label{fig:opponent_conditioned_selection}
\end{figure}

For Fig.~\ref{fig:opponent_conditioned_selection}, each panel uses one controlled observation and 500 independent categorical samples. Panels (a,b) vary opponent lateral position with previous target centered; (c,d) vary previous target with opponent centered. Ball history, self position, and all other inputs are fixed; each cell reports empirical selection frequency.

\begin{figure*}[!t]
    \centering
    \includegraphics[width=\textwidth,pagebox=cropbox]{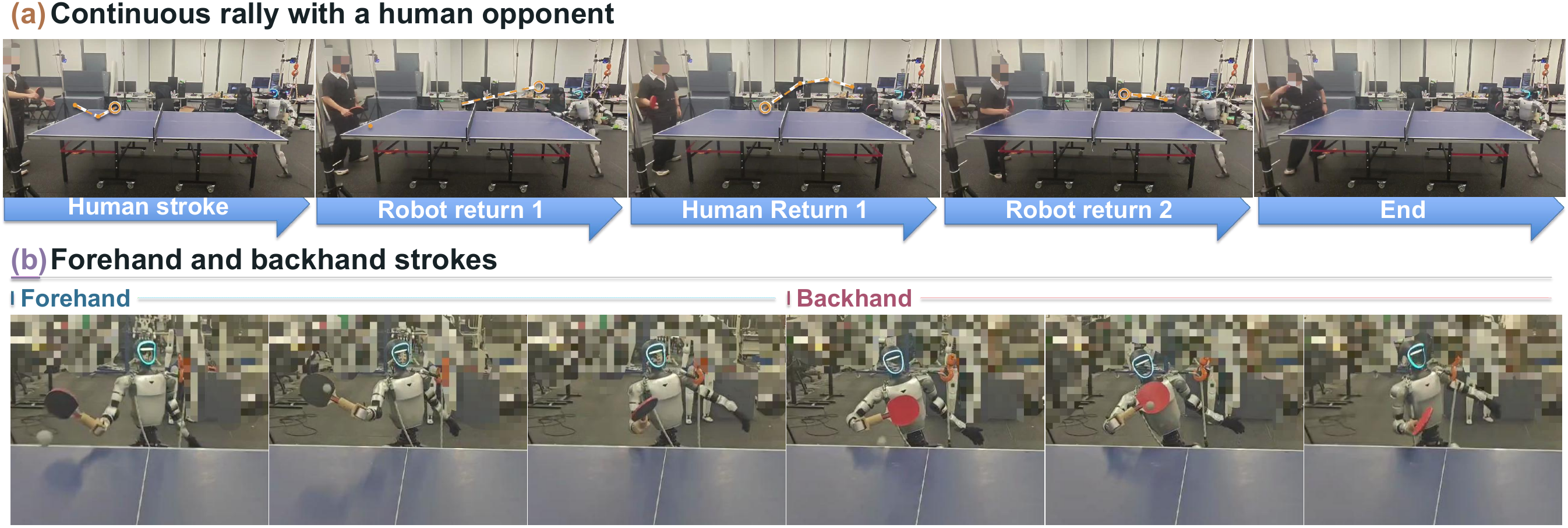}
    \caption{Real-world deployment of the complete hierarchy in human--robot play. (a) A multi-stroke rally containing two robot returns and one intervening human return. (b) Representative forehand and backhand strokes. The same frozen execution stack and landing-target interface are retained during deployment. Complete videos of the three reported 11-point games are provided in the supplementary material.}
    \label{fig:real_robot_snapshots}
\end{figure*}

\subsection{RQ1: How Do Landings Follow Commanded Targets?}

In simulation, 99.2\% of returns landed legally; of these, 94.3\% were within 30~cm of the commanded target. The pooled landing error across nine targets was $15.0 \pm 8.4$~cm (mean $\pm$ sample standard deviation). Fig.~\ref{fig:target_landing_control}(a) shows that achieved landing position varies with the continuous command. Fig.~\ref{fig:target_landing_control}(b) shows target-conditioned hardware execution over legal first opponent-half landings. Together, they demonstrate target-conditioned execution with nonuniform error across targets.

\subsection{RQ2: Does Interaction-Driven Training Improve Rally Execution Beyond Additional Training?}

Under matched evaluation conditions, we compare the original Stage-I checkpoint with two continuation branches initialized from it and trained with the same additional budget. Prescribed continuation retains the Stage-I incoming-ball distribution, whereas interaction-driven self-play uses rally-generated incoming states. In Fig.~\ref{fig:target_landing_control}(c), interaction-driven self-play achieves higher target-landing accuracy and legal-return rate than both Stage I and the matched-budget prescribed continuation across the displayed rally indices; Fig.~\ref{fig:launch_state_distribution} shows distinct spatial launch patterns generated by prescribed balls and racket returns.

\subsection{RQ3: Competitive Outcomes Under Matched Execution}

Under matched execution, the learned selector achieves $88.9\pm1.4$\%, $81.9\pm2.0$\%, and $73.1\pm4.3$\% point win rates against Center, Random, and Away, respectively (mean $\pm$ sample standard deviation across eight evaluation seeds; Fig.~\ref{fig:pairwise_win_rate}). This includes the opponent-position-aware Away rule, which selects the region farthest from the opponent. All selectors share the frozen execution stack, so the comparison changes target selection, not learned motor parameters.
\subsection{RQ4: Opponent-Position and Previous-Target Sensitivity}

Changing opponent lateral position alone shifts selection approximately opposite the opponent (Fig.~\ref{fig:opponent_conditioned_selection}(a,b)); changing previous landing target alone shifts next-target selection approximately opposite that target (Fig.~\ref{fig:opponent_conditioned_selection}(c,d)). These controlled interventions establish local sensitivity to opponent position and previous target, without attributing the competitive advantage to either input individually.

\subsection{Real-World Human--Robot Matches}

During physical play against a recreational human player, the motor actor $\pi_L$ and its observation normalizer remain frozen, with the strike planner and downstream joint controller unchanged. The learned $\pi_H$ receives the same strategic state, replacing simulated opponent position with motion-captured human planar position. At each robot stroke event, $\pi_H$ selects one of nine landing regions; the same continuous landing-target interface supplies the command to the planner and frozen motor actor.

Across three reported 11-point games with the same recreational human opponent, the recorded scores were 11--4, 11--6, and 11--9 in favor of the robot. The games include multi-stroke rallies and both forehand and backhand returns (Fig.~\ref{fig:real_robot_snapshots}). These games demonstrate the complete hierarchy with the same strategic-to-motor interface and frozen execution stack.

\section{Discussion and Limitations}
\label{sec:discussion}

\subsection{Why Separate Motor and Strategy Learning?}

Stage~II changes execution experience while preserving target semantics; Stage~III changes target selection over frozen execution. More broadly, this suggests a reusable design principle: keep high-level task outcomes semantically fixed while low-level execution adapts. Physical games retain this boundary.

\subsection{Current Limitations}

External ball tracking and motion capture simplify perception. The ball model omits spin and Magnus forces. Nine landing regions limit strategic resolution, while planar opponent position omits pose, motion intent, and identity; the policy therefore has no explicit persistent opponent model. Serving uses a fixed target. Real-world evaluation currently covers one recreational opponent; broader competitive generalization remains to be assessed across additional players, skill levels, and playing styles.

\section{Conclusion}
\label{sec:conclusion}

We presented a hierarchical humanoid table-tennis system with a persistent landing-target command separating whole-body execution from strategic selection. Interaction-driven training adapts the target-conditioned motor layer to closed-loop rallies, then freezes the execution stack for strategic self-play. Under matched execution, the learned selector achieves higher simulated point win rates than Center, Random, and opponent-aware Away, while the hierarchy supports complete real-world human--robot games. The interface is both a control and experimental boundary between motor adaptation and strategic learning, connecting a target-conditioned hitter to a strategic player.

\bibliographystyle{IEEEtran}
\bibliography{references}

\end{document}